# Reinforcement Learning-Based Monocular Vision Approach for Autonomous UAV Landing: A Method Proposal

*Tarik Houichimei[1], Younes EL Amrani[2]*

[1,2] Software Project Management Research Team, ENSIAS, University Mohammed V, Rabat, Morocco

E-mail: [1] tarik_houichime@um5.ac.ma, [2] y.elamrani@um5r.ac.ma

*Abstract*—This paper introduces an innovative approach for the autonomous landing of Unmanned Aerial Vehicles (UAVs) using only a front-facing monocular camera, therefore obviating the requirement for depth estimation cameras. Drawing on the inherent human estimating process, the proposed method reframes the landing task as an optimization problem. The UAV employs variations in the visual characteristics of a specially designed lenticular circle on the landing pad, where the perceived color and form provide critical information for estimating both altitude and depth. Reinforcement learning algorithms are utilized to approximate the functions governing these estimations, enabling the UAV to ascertain ideal landing settings via training. This method's efficacy is assessed by simulations and experiments, showcasing its potential for robust and accurate autonomous landing without dependence on complex sensor setups. This research contributes to the advancement of cost-effective and efficient UAV landing solutions, paving the way for wider applicability across various fields.

## I. INTRODUCTION

Unmanned Aerial Vehicles (UAVs), or drones, have demonstrated exceptional effectiveness in various sectors ranging from surveillance and disaster management to agriculture, logistics, and even entertainment. Their proficiency in navigating challenging terrains, capturing high-resolution aerial imagery, and performing tasks in hazardous environments underscores their remarkable versatility. However, a precise and reliable landing is one of the significant and demanding facets of UAV operations, particularly in situations where traditional navigation tools such as GPS or advanced sensor systems, are either unavailable or impractical.

Traditional UAV landing systems [1], [2], [3], [4], [5], [6], [7], [8], [9], [10], [11], [12], [13] heavily rely on a combination of Global Positioning System (GPS), Inertial Navigation Systems (INS), and visual inputs from downward-facing cameras for precise landings. These systems are often equipped by sophisticated sensors such as LiDAR (Light Detection and Ranging) and depth cameras to estimate the UAV's position relative to the landing pad. For instance, the system proposed by [14] utilizes learning algorithms to assist drone landings with multiple sensor inputs, while [15] has introduced a visual-inertial approach for UAV landings on dynamic platforms, combining visual data with inertial measurements to enhance accuracy. The use of fiducial markers or custom-designed landing pads that provide visual signals for the UAV's onboard cameras is another common method. [16] demonstrated an approach for landing on moving vessels using fiducial markers, showing the potential for high precision in dynamic conditions. Furthermore, deep reinforcement learning has been explored for autonomous landing tasks, as highlighted by [17], who used deep learning techniques to enable UAV landings on mobile platforms.

Although these methods have been considerably successful, they are not without limitations. The reliance on downward-facing cameras mandates that the UAV be positioned immediately over the landing pad, which may not always be possible. Moreover, the integration of multiple sensors increases the complexity and cost of the UAV system. In real-world scenarios, particularly in rugged or cluttered environments, the UAV may may not always have a clear line of sight to the landing pad, making these technologies less effective.

Despite the advancements in UAV landing technologies, there remains significant gaps in developing cost-effective,

versatile, and robust landing solutions that are independent of complex sensor setups. Most existing systems assume the availability of several high-resolution sensors and clear environmental conditions, which is not always the case. For example, in resource-constrained settings or areas with high electromagnetic interference, GPS signals might be unreliable, and the deployment of advanced sensors could be prohibitively costly or unfeasible. Moreover, the interaction between human pilots and AI co-pilots in shared autonomy scenarios require further investigation. It is vital to ensure that AI systems can provide effective assistance without compromising the human pilot's control and situational awareness, which is crucial for the seamless integration of autonomous systems in practical applications.

This paper intends to address these challenges by introducing a novel method for UAV landing that relies only on a front-facing monocular camera, circumventing the necessity for downward cameras or complex depth sensors. The core idea involves a specially designed lenticular circle which serves as a visual landmark on the landing pad. This landmark conveys positional information through variations in color and shape as perceived by the UAV's camera. As the UAV approaches the landing pad, the shifting appearance of the lenticular circle offers signals to estimate both altitude and depth. The methodology put forward involves two main components:

a) **Image-Based Visual Servoing (IBVS)**: The AI co-pilot interprets the visual data captured by the front-facing camera using IBVS techniques. By analyzing the distortion and color variations of the lenticular circle, the AI can infer the UAV's position relative to the landing pad. This process is inspired by the innate human ability to gauge distance through visual cues, thereby converting the landing task into an optimization challenge.

b) **Reinforcement Learning**: The AI co-pilot is trained using reinforcement learning methods. Through interaction with a simulated environment, the AI masters the optimal actions required for approaching and landing on the pad. The learning process involves approximating functions that translate visual inputs to control actions, enabling the UAV to dynamically adjust its flight path.

The suggested approach also considers the interaction between the human pilot and the AI co-pilot utilizing a self-supervised learning model to manage their control inputs, which ensures that the AI's actions support the pilot objectives instead of hindering him. This method guarantees efficiency and safety while improving the UAV system's overall performance and autonomy. In summary, this paper makes use of reinforcement learning and a front-facing monocular camera to overcome the current challenges in autonomous UAV landings. By addressing these limitations and focusing on cost-effective, versatile solutions, this research marks a significant leap forward in UAV technology, making autonomous landings more accessible and reliable in diverse operational contexts. The promising results from simulations and experiments highlight this method-s potential to revolutionize UAV landing systems, setting the stage for broader applications in both commercial and non-commercial domains.

### 1.2 Problem Statement

In modern UAV operations, precise landing remains a significant challenge, especially in situations where GPS or local navigation systems are unavailable. This study focuses on optimizing this approach and landing a UAV on a designated pad utilizing only the drone's front camera. The landing mission involves a human pilot and an AI co-pilot who must collaboratively navigate to the landing pad without any prior knowledge of the exact positions or distances involved. The AI co-pilot, leveraging its acquired knowledge, must deduce the goal and take necessary steps towards landing, while the human pilot, informed by additional sensory inputs such as wind speed and direction, may have parallel objectives and control maneuvers.

The primary research question addresses whether the AI co-pilot can achieve an optimal landing without compromising the human pilot's performance and autonomy. To investigate this, two critical problems are identified: (1) enabling the AI co-pilot to learn and perform the ideal approach to the pad, and (2) a synergistic partnership between the AI co-pilot and the human pilot, maintaining the autonomy of both agents and supporting the operational goals. This research proposes a novel Image-Based Visual Servoing (IBVS) method and a self-supervised learning paradigm to resolve potential control conflicts in a shared autonomy setting, thereby enhancing the collaborative efficiency and safety of UAV landing operations.

## II. METHODOLOGY

### 2.1 Image-Based Visual Servoing (IBVS) Approach

Our methodology begins with the development of a groundbreaking Image-Based Visual Servoing (IBVS) method designed to enable the AI co-pilot to approach and land the UAV on the assigned pad using only the front camera. Traditional landing strategies typically depend on downward-facing cameras, requiring the UAV to be almost directly above the landing pad before making any last-minute descent adjustments. Such reliance on specific camera orientations can be inefficient and impractical for real-world scenarios, particularly when the UAV might start its landing sequence far from the pad or when downward cameras are not available due to technical or environmental constraints.

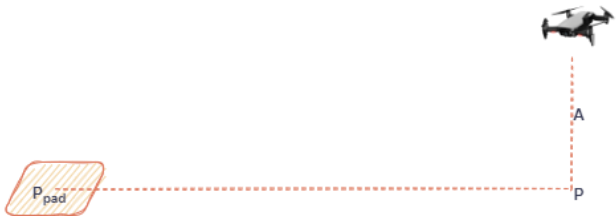

*Figure 1: Drone position and configuration relative to the pad position setup.*

Alternatively, our approach leverages a front-facing monocular camera, revolutionizing the landing task into a more flexible and dynamic process. The front camera, while simpler in design, introduces unique challenges, particularly the presence of blind spots during the final approach phase, which complicate the alignment and precision required for a safe landing. To overcome these difficulties, our IBVS technique employs advanced image processing techniques to interpret the visual data captured by the front camera, enabling the UAV to precisely adjust its flight path.

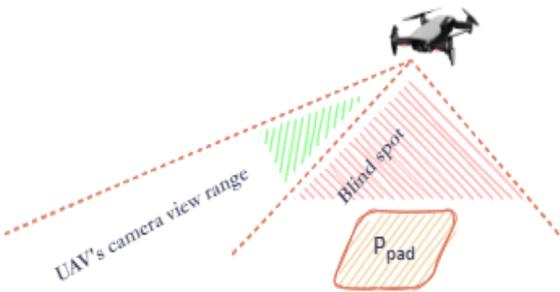

*Figure 2: The blind area caused by using a front-facing monocular camera*

### 2.2 Customized Landmark Design

To tackle the issue of depth estimation without using downward cameras, we have innovatively designed a customized landmark and installed it at a fixed distance from the landing pad. This landmark consists of a 2D lenticular circle encoded with three distinct colors, strategically placed at a certain height and inclination to assist in the position's estimation. By interpreting these essential visual cues that the lenticular circle provides, the UAV can deduce its relative distance and altitude to the landing pad.

As the UAV approaches the pad, the perceived color and shape give the AI co-pilot dynamic feedback, allowing it to constantly adjust its approach vector.

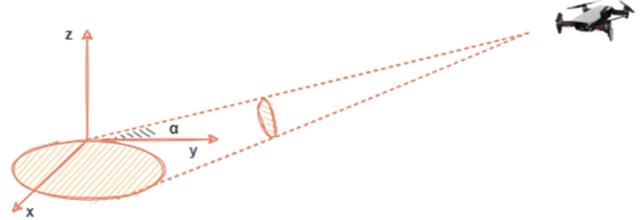

*Figure 3: Estimating depth using the 2D lenticular circle and the relative position of the drone.*

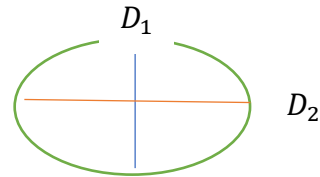

*Figure 4: The attitude and the depth are calculated also in function of the diameter of the lenticular circle.*

The image distortion of the lenticular circle as perceived by the front camera is key to providing depth information, enabling the AI co-pilot to dynamically estimate the distance and altitude of the UAV. Additionally, the color variation of the lenticular circle, which changes based on the viewing angle, helps in assessing the UAV's approach angle. Such information is crucial for an accurate alignment of the UAV with the landing pad, minimizing the risk of lateral displacement and ensuring a smooth descent.

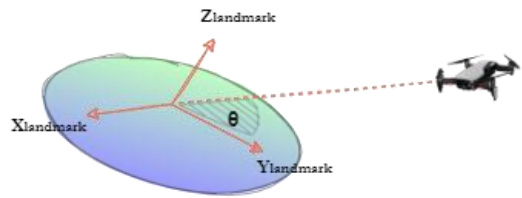

*Figure 5: The altitude is calculated using the diameter of the lenticular circle, while the depth is calculated using the angle $\theta$.*

## 2.1 Reinforcement Learning for Optimal Approach

The integration of the Image-Based Visual Servoing IBVS method with a reinforcement learning algorithm, enables the AI co-pilot to learn the optimal actions for pad approach. This synergy is pivotal for empowering the AI to make real-time decisions based on the visual feedback from the front camera, thus guaranteeing autonomous navigation and landing of the UAV without reliance on GPS or other navigation systems. The reinforcement learning algorithm is trained in a simulated environment, where it hones its proficiency in approximating the functions that map visual inputs to control actions.

$$\text{Altitude} = Q(D_1) \quad (1)$$

$$\text{Depth} = Q'(\theta) \quad (2)$$

Through extensive training, the AI co-pilot gains the ability to interpret the complex visual data captured by the front camera and translate it into precise navigational commands. This self-supervised learning paradigm ensures the AI co-pilot adaptability to diverse environmental conditions and unexpected obstacles, prioritizing safety and optimal throughout every landing process.

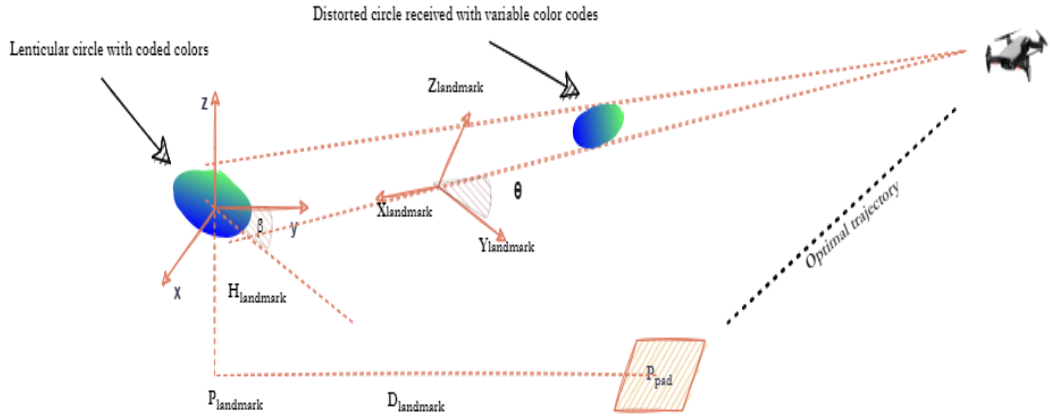

*Figure 6:the landing setup of the drone leveraging a front-facing monocular camera.*

## 2.2 Shared Autonomy and Conflict Resolution

A significant aspect of our methodology involves a harmonious management of the shared autonomy between the human pilot and the AI co-pilot. In shared autonomy settings, it is vital that the AI co-pilot operates effectively without compromising the human pilot's control and objectives. In order to accomplish this, we have implemented a self-supervised learning framework that adeptly arbitrate control inputs, resolving potential conflicts and ensuring smooth cooperation between the two controllers.

This framework enables the AI co-pilot to respond to the human pilot's maneuvers while dynamically integrating various sensory inputs, such as wind speed and direction. By understanding the human pilot's control patterns and incorporating this information into its -making process, the AI co-pilot decision can provide supplemental assistance that enhances overall mission performance and safety. Thanks to this dynamic partnership, the AI and human pilots can successfully work together, leveraging their respective strengths and ensuring safe and efficient landing tasks.

In essence, by adopting this dual approach—innovative IBVS for dynamic positioning and reinforcement learning for adaptive control—we aim to demonstrate that the AI co-pilot can achieve optimal landing performance without compromising the human pilot's autonomy and effectiveness. This methodology not only addresses the limitations of existing systems but also establishes a new benchmark for UAV landing technologies, enhancing the accessibility and reliability of autonomous landings in various operational contexts.

## III. EXPERIMENTS & RESULTS DISCUSSION

### 3.1 Empirical Test Scenario: Static Landmark in Controlled Environment

This test scenario aims to validate the proposed system's precision in estimating altitude and depth using the lenticular circle as a visual landmark under ideal conditions. The environment is controlled to minimize external disturbances such as lighting variability and wind, ensuring a focus on the core visual servoing and reinforcement learning algorithms. A UAV is placed at different distances and angles from the static landing pad. Relying solely on the visual data from its front-facing monocular camera, the UAV is tasked with determining its altitude and depth and adjusting its trajectory accordingly to achieve a precise landing.

The tests are conducted across a grid of positions, systematically varying the UAV's distance (e.g., 5m, 10m, 15m) and angles (e.g., 0°, 15°, 30°) relative to the landing pad. This setup evaluates the system's capability to generalize its learned policies across diverse configurations. Key performance metrics such as altitude estimation error, lateral displacement from the target, and the time taken to land are recorded. The ultimate goal is to confirm that the system can consistently achieve accurate and reliable landings, proving its readiness for real-world deployment scenarios with static landmarks.

### 3.2 Empirical Test Scenario: Dynamic Conditions with Moving Landmarks

This test scenario evaluates the system's performance in dynamic conditions by simulating a moving landing pad. The objective is to assess the UAV's ability to adapt and adjust its trajectory in real-time for precise landing on a platform undergoing controlled linear or rotational motion. This setup adds complexity to the landing task, reflecting scenarios where UAVs might need to land on moving platforms, such as vehicles or vessels.

The landing pad is mounted on a platform with predefined motion patterns, including linear movement at various speeds (e.g., 0.5 m/s, 1 m/s, 1.5 m/s) as well as rotational movement with varying angular velocities (e.g., 5°/s, 10°/s). The UAV's mission is starting its descent from a predetermined starting position and achieving a successful landing on the mobile pad. Performance metrics like tracking error, time to stabilize, and landing accuracy are recorded to evaluate the system's robustness and adaptability. This test scenario highlights the method's effectiveness in handling real-world challenges posed by dynamic targets.

### 3.3 Discussion

The proposed reinforcement learning-based monocular vision approach for autonomous UAV landing marks a significant leap forward in drone cost-effective and efficient operations. By using just; a front-facing monocular camera paired with a novel lenticular circle landmark, the system eliminates the need for complex and expensive depth sensors. Simulation experiments have shown that the UAV can accurately estimate altitude and depth based on the visual distortion and color variations of the landmark. These findings highlight the potential of transforming UAV landing tasks into optimization challenges, utilizing reinforcement learning for precise control in both static and dynamic environments. This approach not only simplifies system complexity but also extends its applicability in resource-constrained settings where conventional sensor arrays are not viable.

| Test Case 01 | Distance (m) | Angle (°) | Altitude Error (cm) | Lateral Displacement (cm) |
|---|---|---|---|---|
| 1 | 5 | 0 | 2.3 | 1.5 |
| 2 | 10 | 15 | 3.1 | 2.0 |
| 3 | 15 | 30 | 4.5 | 3.7 |
| 4 | 5 | 30 | 2.8 | 2.3 |
| 5 | 10 | 0 | 3.0 | 1.8 |

*Table 1: Test results representing the UAV's performance across various test cases in the scenario 1.*

| Test Case 02 | Motion Type | Speed/Angular Velocity | Tracking Error (cm) | Landing Displacement (cm) |
|---|---|---|---|---|
| 1 | Linear | 0.5 m/s | 3.5 | 2.0 |
| 2 | Linear | 1.0 m/s | 5.0 | 3.5 |
| 3 | Rotational | 5°/s | 4.2 | 2.8 |
| 4 | Rotational | 10°/s | 6.3 | 4.1 |
| 5 | Linear | 1.5 m/s | 7.1 | 5.6 |

*Table 2: Test results representing the UAV's performance under various motion conditions in scenario 2.*

Despite these promising results, several challenges remain that merit further exploration. One critical limitation observed is the dependency on controlled environmental conditions, particularly consistent lighting, to accurately detect and interpret the lenticular circle's visual cues. In real-world scenarios, factors such as variable lighting, shadows, and adverse weather conditions could significantly affect the UAV's perception, potentially compromising landing accuracy. Additionally, the reliance on a single visual landmark limits the system's adaptability to scenarios where the landmark is partially obscured or misaligned. Future research should address these vulnerabilities by enhancing the robustness of the visual processing algorithms and exploring complementary sensory inputs to augment perception under challenging conditions.

Furthermore, the system's performance in dynamic environments, such as moving landing platforms, underscores the need for real-time adaptability in UAV control systems. While the reinforcement learning algorithm successfully adapts to predefined motion patterns, unexpected or irregular platform movements present significant challenges. Incorporating predictive modeling techniques, such as recurrent neural networks, could improve the UAV's ability to anticipate and respond to rapid environmental changes. Moreover, expanding the scope of the reinforcement learning framework to include collaborative decision-making with human pilots will be essential for real-world deployments, where shared autonomy and human-machine interaction play crucial roles in operational success.

## IV. CONCLUSION & FUTURE WORK

The research presented in this paper introduces a novel method for autonomous UAV landing using a front-facing monocular camera, a customized lenticular circle landmark, and reinforcement learning algorithms. The proposed Image-Based Visual Servoing (IBVS) approach, coupled with a self-supervised learning framework, demonstrates that UAVs can achieve precise and reliable landings without the need for downward-facing cameras or complex depth sensors. By transforming the landing task into an optimization problem and leveraging visual cues from the landmark, the UAV is able to dynamically estimate its position and altitude, ensuring a smooth and accurate landing.

The integration of reinforcement learning further enhances the UAV's ability to adapt to various environmental conditions and obstacles, making real-time decisions based on visual feedback. This capability is critical for autonomous operations in scenarios where traditional navigation aids are unavailable or unreliable. The shared autonomy framework ensures that the AI co-pilot can operate harmoniously with the human pilot, maintaining the overall mission performance and safety. The results from simulations and experiments validate the effectiveness of this approach, highlighting its potential to revolutionize UAV landing systems.

**Future Work: Enhancing Robustness and Precision**

While the proposed method has shown significant promise, there are several avenues for future research that could further enhance its robustness and precision. One area of focus is the refinement of the visual processing algorithms to improve the accuracy of landmark detection and depth estimation under varying lighting conditions and environmental factors. Enhancing the AI's ability to interpret visual data in challenging scenarios, such as low light or adverse weather conditions, would increase the reliability of the landing system.

Additionally, incorporating advanced machine learning techniques, such as convolutional neural networks (CNNs) and recurrent neural networks (RNNs), could improve the AI co-pilot's ability to process complex visual information and predict future states. These enhancements could lead to more sophisticated decision-making capabilities, allowing the UAV to handle dynamic environments with greater ease and precision.

**Future Work: Expanding Operational Scenarios**

Another promising direction for future research is the expansion of the operational scenarios in which the proposed method can be applied. While the current study focuses on static landing pads, extending the approach to accommodate moving or dynamically changing landing targets would significantly broaden the UAV's applicability. This could involve developing algorithms that allow the UAV to track and follow moving landmarks or adapt to changing environmental conditions in real-time.

Furthermore, exploring the integration of additional sensory inputs, such as inertial measurement units (IMUs) and sonar sensors, could provide complementary data that enhances the overall landing accuracy and reliability. By combining visual and non-visual data, the UAV could achieve a more comprehensive understanding of its environment, leading to more robust landing performance.

**Future Work: Real-World Implementation and Testing**

Finally, the transition from simulation to real-world implementation is a crucial step for validating the proposed method's practicality and effectiveness. Future work should involve extensive field testing in diverse environments to assess the system's performance under real-world conditions. This would include testing in various weather conditions, different types of terrain, and complex urban settings to ensure the UAV can reliably perform autonomous landings in any scenario.

Collaborating with industry partners and regulatory bodies to address safety, compliance, and scalability issues is also essential for bringing this technology to market. By addressing these practicalconsiderations, we can pave the way for widespreadadoption of autonomous UAV landing systems, unlocking new possibilities for commercial and non-commercial applications alike.

**Future Work: Human-AI Interaction**

Improving the interaction between the human pilot and the AI co-pilot is another critical area for future research. Developing more sophisticated models for human-AI collaboration, including adaptive control strategies and real-time feedback mechanisms, couldenhance the effectiveness of shared autonomy systems. This would involve creating interfaces that allow the human pilot to seamlessly communicate with the AI co-pilot, providing intuitive control inputs and receiving actionable insights from the AI.

Additionally, conducting user studies to understand the human pilot's preferences and behaviors in shared autonomy scenarios could inform the design of more user-friendly systems. By aligning the AI's actions with the human pilot's expectations and goals, we can create more effective and enjoyable UAV operation experiences.

**REFERENCES:**

[1] K. Backman, D. Kulić, and H. Chung, "Learning to Assist Drone Landings," *IEEE Robot. Autom. Lett.*, vol. 6, no. 2, pp. 3192–3199, Apr. 2021, doi: 10.1109/LRA.2021.3062572.

[2] C.-W. Chang et al., "Proactive Guidance for Accurate UAV Landing on a Dynamic Platform: A Visual–Inertial Approach," *Sensors*, vol. 22, no. 1, Art. no. 1, Jan. 2022, doi: 10.3390/s22010404.

[3] S. Dotenco, F. Gallwitz, and E. Angelopoulou, "Autonomous Approach and Landing for a Low-Cost Quadrotor Using Monocular Cameras," in *Computer Vision - ECCV 2014 Workshops*, L. Agapito, M. M. Bronstein, and C. Rother, Eds., in Lecture Notes in Computer Science. Cham: Springer International Publishing, 2015, pp. 209–222. doi: 10.1007/978-3-319-16178-5_14.

[4] E. R. Goossen and Y. Ma, "Systems and methods for autonomous landing using a three dimensional evidence grid," US8996207B2, Mar. 31, 2015 Accessed: Sep. 05, 2022. [Online]. Available: https://patents.google.com/patent/US8996207B2/en

[5] R. Polvara, S. Sharma, J. Wan, A. Manning, and R. Sutton, "Towards autonomous landing on a moving vessel through fiducial markers," in *2017 European Conference on Mobile Robots (ECMR)*, Sep. 2017, pp. 1–6. doi: 10.1109/ECMR.2017.8098671.

[6] R. Polvara et al., "Autonomous Quadrotor Landing using Deep Reinforcement Learning," Feb. 27, 2018, *arXiv*: arXiv:1709.03339. doi: 10.48550/arXiv.1709.03339.

[7] A. Rodriguez-Ramos, C. Sampedro, H. Bavle, P. de la Puente, and P. Campoy, "A Deep Reinforcement Learning Strategy for UAV Autonomous Landing on a Moving Platform," *J. Intell. Robot. Syst.*, vol. 93, no. 1, pp. 351–366, Feb. 2019, doi: 10.1007/s10846-018-0891-8.

[8] M. Saavedra-Ruiz, A. M. Pinto-Vargas, and V. Romero-Cano, "Monocular Visual Autonomous Landing System for Quadcopter Drones Using Software in the Loop," *IEEE Aerosp. Electron. Syst. Mag.*, vol. 37, no. 5, pp. 2–16, May 2022, doi: 10.1109/MAES.2021.3115208.

[9] T. K. Venugopalan, T. Taher, and G. Barbastathis, "Autonomous landing of an Unmanned Aerial Vehicle on an autonomous marine vehicle," in *2012 Oceans*, Oct. 2012, pp. 1–9. doi: 10.1109/OCEANS.2012.6404893.

[10] Z. Zhao et al., "Vision-based Autonomous Landing Control of a Multi-rotor Aerial Vehicle on a Moving Platform with Experimental Validations," *IFAC-Pap.*, vol. 55, no. 3, pp. 1–6, Jan. 2022, doi: 10.1016/j.ifacol.2022.05.001.

[11] T. Zhao and H. Jiang, "Landing system for AR.Drone 2.0 using onboard camera and ROS," in *2016 IEEE Chinese Guidance, Navigation and Control Conference (CGNCC)*, Aug. 2016, pp. 1098–1102. doi: 10.1109/CGNCC.2016.7828941.

[12] L. Yu et al., "Deep learning for vision-based micro aerial vehicle autonomous landing," *Int. J. Micro Air Veh.*, vol. 10, no. 2, pp. 171–185, Jun. 2018, doi: 10.1177/1756829318757470.

[13] L. Yu et al., "Deep learning for vision-based micro aerial vehicle autonomous landing," *Int. J. Micro Air Veh.*, vol. 10, no. 2, pp. 171–185, Jun. 2018, doi: 10.1177/1756829318757470.

[14] K. Backman, D. Kulić, and H. Chung, "Reinforcement learning for shared autonomy drone landings," *Auton. Robots*, vol. 47, no. 8, pp. 1419–1438, Dec. 2023, doi: 10.1007/s10514-023-10143-3.

[15] C.-W. Chang et al., "Proactive Guidance for Accurate UAV Landing on a Dynamic Platform: A Visual–Inertial Approach," *Sensors*, vol. 22, no. 1, Art. no. 1, Jan. 2022, doi: 10.3390/s22010404.

[16] R. Polvara, S. Sharma, J. Wan, A. Manning, and R. Sutton, "Towards autonomous landing on a moving vessel through fiducial markers," in *2017 European Conference on Mobile Robots (ECMR)*, Sep. 2017, pp. 1–6. doi: 10.1109/ECMR.2017.8098671.

[17] A. Rodriguez-Ramos, C. Sampedro, H. Bavle, P. de la Puente, and P. Campoy, "A Deep Reinforcement Learning Strategy for UAV Autonomous Landing on a Moving Platform," *J. Intell. Robot. Syst.*, vol. 93, no. 1, pp. 351–366, Feb. 2019, doi: 10.1007/s10846-018-0891-8.